\documentstyle[12pt,A490,epsbox]{article}
\input{standard92.mac}
\input{local.mac}

\begin{document}

\begin{tabbing}
{\bf Technical Reports on Mathematical and Computing Sciences:} TR-C124\\
{\bf title:}~
A role of constraint in self-organization\\
{\bf authors:}~
Carlos Domingo$^1$, Osamu Watanabe$^2$, and Tadashi Yamazaki$^2$\\
{\bf affiliation:}\\
1.~\=\kill
1.\>Dept.\ de LSI,
    Univ.\ Politecnica de Catalunya\\
  \>Campus Nord, Modul C5, 08034-Barcelona, Spain.\\
  \>{\bf email:} carlos@lsi.upc.es\\
2.\>Dept.\ of Mathematical and Computing Sciences,
    Tokyo Institute of Technology\\
  \>Meguro-ku Ookayama, Tokyo 152-8552.\\
  \>{\bf email:} $\{$watanabe, tyam$\}$@is.titech.ac.jp\\
{\bf acknowledgements to financial supports:}\\
1.\>Supported in part by ESPRIT LTR Project no.\ 20244 - ALCOM-IT\\
  \>and CICYT Project TIC97-1475-CE.\\
2.\>Supported in part by the Ministry of Education, Scinece, Sports
    and Culture,\\
  \>Grant-in-Aid for Exploratory Research, ????, 1997.
\end{tabbing}

\noindent
{\bf Abstract.}~
In this paper we introduce a neural network model of
self-organization.  This model uses a variation of Hebb rule for
updating its synaptic weights, and surely converges to the equilibrium
status.  The key point of the convergence is the update rule that
constrains the total synaptic weight and this seems to make the model
stable. We investigate the role of the constraint and show that it is
the constraint that makes the model stable. For analyzing this
setting, we propose a simple probabilistic game that models the neural
network and the self-organization process. Then, we investigate the
characteristics of this game, namely, the probability that the game
becomes stable and the number of the steps it takes.

\smallsection{Introduction}

How does the brain establish connections between neurons?  This
question has been one of the important issues in Neuroscience, and
theoretical researchers have proposed various models for
self-organization mechanisms of the brain.  In many of these models,
competitive learning, or more specifically, competitive variants of
Hebb's rule have been used as a key principle.  In this paper, we
study one property of such competitive Hebb rules.

As one typical example of self-organization, ``orientation
selectivity'' \cite{wiesel-hubel:JNeuro63} has been studied
intensively.  In the primary visual cortex (area17) of cats, there is
some group of neurons that strongly reacts to the presentation of
light bars of a certain orientation, which we call {\em orientation
selectivity}.  An interesting point is that in a very early stage
after birth, every neuron reacts to all bars of every orientation.
This indicates that orientation selectivity is obtained after birth;
that is, each neuron selects one preferred orientation among all
orientations.  To explain the development of orientation selectivity,
a considerable number of mathematical models have been investigated;
see, e.g., \cite{swindale96:Network96}.  Although these models may
look quite different, most of them use, as a principal rule for
modifying synaptic strength, a competitive variant of Hebb rule, which
is essentially the same as the rule proposed in the pioneer paper of
von der Malsburg \cite{malsburg:kybernetik73}, the paper that first
gave a mathematical model for the development of orientation
selectivity.

A {\em Hebb rule} is a simple rule for updating, e.g., the weight of
connection between two neurons.  The rule just says that the
connection between two neurons is strengthened if they both become
active simultaneously.  This rule has been used widely for neural
network learning.  Von der Malsburg constrained this updating rule so
that the total connection weight of one neuron are kept under some
bound.  In this paper, we call this variation of Hebb rule a {\em
constrained Hebb rule}.  He showed through computer experiments that
orientation selectivity is surely developed with his constrained Hebb
rule.

Since the work of von der Malsburg, many models have been proposed,
and some have been theoretically analyzed in depth; see, e.g.,
\cite{tanaka:NeuralNetwork90}.  For example, a feature of various
constrained Hebb rules as a learning mechanism has been discussed in
\cite{miller-mackay:NeuralComp94}.  Yet, the question of 
 why orientation selectivity is obtained by following a constrained
Hebb rule has not been addressed.  Note that the development of
orientation selectivity is different from ordinary learning in the
sense that a neuron (or, a group of neurons) establishes a preference
to one particular orientation from given (more or less) uniformly
random orientation stimuli.  In this paper, we discuss why and how one
feature from equally good features is selected with a constrained Hebb
rule.

In order to simplify our analysis, we propose a simple probabilistic
game called ``monopolist game'' for abstracting Hebb rules.  In
monopolist game, an updating rule corresponds to game's rule, and the
selectivity is interpreted as that a single winner of a game ---
monopolist --- emerges.  Then we prove that a monopolist emerges with
probability one in games following a von der Malsburg type rule.  On
the other hand, we showed theoretical evidence supporting that (i) the
chance of having a monopolist is low without any constraint, and (ii)
a monopolist emerges even under a rule with a weaker constraint.
These results indicates that the importance of constraint in Hebb
rules (or, more generally, competition in learning) to select one
feature from equally good features.

We also analyzed how fast the monopolist emerges in games following a
von der Malsburg type rule.  This analysis can be used, in future, to
estimate the convergence speed of constrained Hebb rules.
(In this paper,
most of the proofs are given in Appendix.)

\smallsection{Von der Malsburg's Model and Monopolist Game}

Here we first explain briefly
the model considered by von der Malsburg.
(Von der Malsburg studied the selectivity for a set of neurons,
but here we only consider its basic component.)

\medskip\noindent
\underline{Neural Network Structure}

\noindent
We consider two layer neural network.
In particular,
we discuss here the orientation selectivity for one neuron,
and thus,
we assume that there is only one output cell.
On the other hand,
the input layer consists of 19 input cells
that are (supposed to be) arranged in a hexagon like the ones in Figure~1.
We use $i$ for indicating the $i$th input cell,
and $\mbox{\it IN}$ for the set of all input cells.

\medskip\noindent
\underline{Stimuli and Firing Rule}

\noindent
We use 9 stimuli with different orientations (Figure 2),
which are given to the network randomly.
Here $\bullet$ indicates an input cell that gets input 1,
and $\circ$ indicates
an input cell that gets input 0.

\displayskip
\begin{center}
\begin{minipage}{0.44\textwidth}
\psbox[width=\textwidth]{fig1.ps}~
\displayskip
\hfil
Figure~1: Neural network model.
\hfil
\end{minipage}
~~~
\begin{minipage}{0.5\textwidth}
\psbox[width=0.9\textwidth]{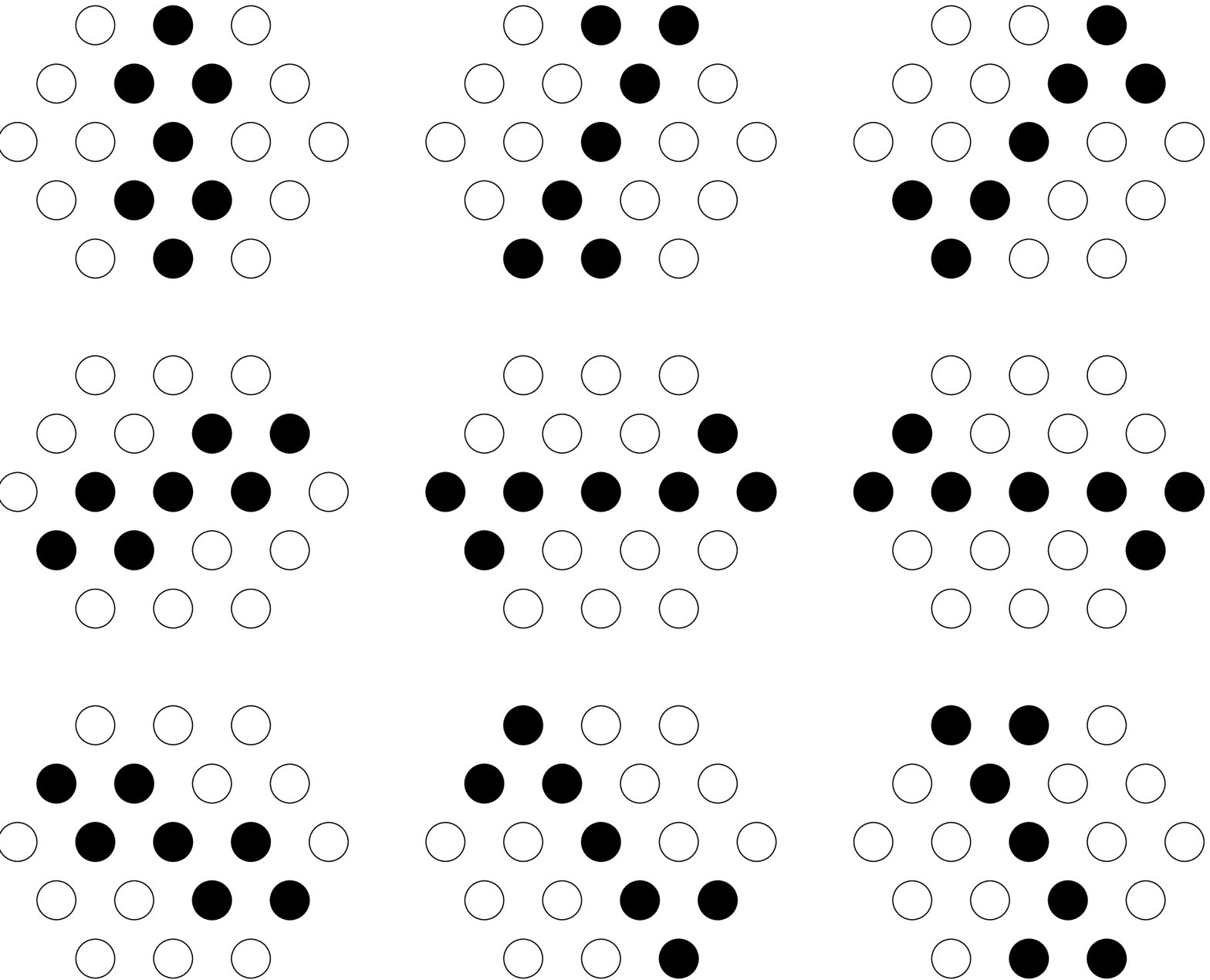}~
\displayskip
\hfil
Figure~2: Nine stimuli.
\hfil
\end{minipage}
\end{center}
\displayskip

We use $a_i$ to denote input value (either 0 or 1) to the $i$th input cell.
Then output value $V$ is computed as
$V$ $=$
$\mbox{\rm Th}_p\left(\sum_{i\in{\it IN}}w_i a_i\right)$,
where $w_i$ is the current synaptic strength
between the output cell and the $i$th input cell.
$\mbox{\rm Th}_p(x)$ is a threshold function
that gives $x-p$ if $x>p$ and 0 otherwise,
where $p$ is given as a parameter.

\medskip\noindent
\underline{Updating Rule}

\noindent
Initially,
each weight $w_i$ is set to some random number between 0 to some constant.
Then weights are updated each time
according to the following variation of Hebb's rule,
which we call the {\em constrained Hebb rule} (of von der Malsburg).

\OMIT{\[
\begin{array}{lcl}
w'_i&=&w_i+\cinc a_iV,{\rm~~and}
\displaystyle
w_i &=&w'_i\times\left(W_0/\sum_{k\in I}w'_{k}\right).
\end{array}
\]}

\[
w'_i~=~w_i+\cinc a_iV,{\rm~~and~~}
w_i ~=~w'_i\times\left(W_0/\sum_{k\in I}w'_{k}\right).
\]

\noindent
Where $\cinc$ (which is called a {\it growth rate})
and $W_0$ ({\it total weight bound}) are constants given as parameters.
The first formula may be considered as the original Hebb's rule;
on the other hand,
the second one is introduced
in order to keep the total weight within $W_0$.
(In fact,
it is kept as $W_0$.)

\medskip
With this setting,
von der Malsburg demonstrated
that the selectivity is developed through computer simulations.
Thus,
it seems likely that
some selection occurs
even from uniformly random examples,
and that the constraint of the von der Malsburg's rule is
a key for such a selection.
In this paper
we would like to study this feature of the constrained Hebb rule.
For this,
we further simplify von der Malsburg's computation model,
and propose the following simple probabilistic game.

\displayskip\noindent
\hskip15pt
{\bf Monopolist Game}

\noindent\hskip15pt\hangindent15pt
\underline{Basic Rule}:~
Consider a finite number of players.
Initially they are given the same amount of money.
The game goes step by step,
and at each step,
one of the players wins with the same probability.
The winner gets some amount of money,
while the other loses some.

\noindent\hskip15pt\hangindent15pt
\underline{Details}:~
A player who loses all his money is said become {\em bankrupt}.
Once a player becomes bankrupt,
he cannot get any amount of money,
though he can still win with the same probability.
(See below for the motivation.)

\noindent\hskip15pt\hangindent15pt
\underline{Goal}:~~
The game terminates if all but one player become bankrupt.
If the survived player keeps enough money at that point,
then he is called a {\it monopolist}.
We call a situation where a monopolist appears {\it monopoly}.

\beginsome{Notations.}
We use $n$ and $n'$ to denote
the number of players and
that of remaining (not being bankrupt) players,
and use $i$, $1\le i\le n$, to denote players' indices.
Throughout this paper,
each player's wealth is simply called a {\em weight},
and let $w_i$ denote the player $i$'s current weight.
Let $I$ and $W_0$ respectively denote
the initial weight of each player and the total amount of initial weights;
that is,
$W_0=nI$.
\endsome

The connection of this game
with von der Malsburg's computation model is clear;
each player's weight corresponds to
total synaptic strength
between the output cell and a set of input cells
corresponding to one type of stimulus,
and the emergence of a monopolist means that
the network develops preference to one orientation.
From this correspondence,
it is natural to require that
even a bankrupt player can win with the same probability $1/n$,
which reflects the fact that
the probability of a stimulus of each orientation appears is the same
no matter how neural connections are organized.

An updating rule of players' weights corresponds to
a rule of changing synaptic strength in the network.
Here we can state updating rules in the following way.
(In the following,
let $i_0$ denote the player who wins at the current step.)

\[
w_i~=~
\left\{
\begin{array}{ll}
w_i+\finc-\fdec,&\mbox{if $i=i_0$, and}\cr
w_i-\fdec,       &\mbox{otherwise.}
\end{array}
\right.
\]

Here $\finc$ and $\fdec$
are the amount of increment and decrement at each step respectively,
and one type of monopolist game is specified
by defining $\finc$ and $\fdec$.
In the following,
we assume that
these values are determined
from $w_i$, $w_{i_0}$, $n$, and $n'$.
From the relation to von der Malsburg's computation model,
we require that
both $\finc$ and $\fdec$ are 0 if $w_i=0$;
that is,
once a player loses all money,
he stays forever in the 0 weight state.
(In the following,
we will omit stating this requirement explicitly.)

Now we consider the rule that corresponds
to the constrained Hebb rule of von der Malsburg's rule.
It is defined as follows with constant $\cinc$.
\begin{equation}
\finc~=~\cinc,{\rm~~and~~}\fdec~=~\cinc/n'.
\end{equation}
(Recall that $n'$ is the number of currently remaining players.)

Note that
with this rule,
the total amount of wealth is kept constant.
Thus,
in this sense,
it corresponds to von der Malsburg's rule,
and we call it {\em constrained rule}.
Note that
we may also consider a similar rule
such that $\finc$ is not constant
but proportional to $w_i$.
(Similarly,
$\fdec$ is also proportional to $w_i$.)
This rule might be closer to the original von der Malsburg's rule.
This difference is, however, not essential
for discussing the probability of having a monopolist,
i.e., for our discussion in Section 3.
On the other hand,
there is a significant difference in convergence speed;
but roughly speaking,
the difference disappears if we take the log of weight.
Thus,
we will discuss with the above simpler rule.

\smallsection{Importance of Competition}

Here we compare three different updating rules for monopolist game,
and show that
constraint is important to derive a monopolist.
From this,
we could infer that
some sort of constraint,
(or, competition in more general)
is important in learning rules
for selecting one among the others through random process.

In the following,
we consider the following three updating rules:
(1) constrained rule,
(2) local rule, and
(3) semi local rule.
Below we define these rules
(except (1) that has been defined in the previous section)
and discuss
the probability $P_*$ that a monopolist emerges.

\medskip\noindent
\underline{Constrained Rule}

\noindent
We show that
under constrained rule,
$P_*$ is 1,
that is,
a monopolist emerges with probability 1.

A monopolist game in general expressed by an one-dimensional random walk.
More precisely,
for any $i$,
we can express the player $i$'s wealth $w_i$ as the following random walk.

\displayskip
\begin{center}
\psbox[width=0.6\textwidth]{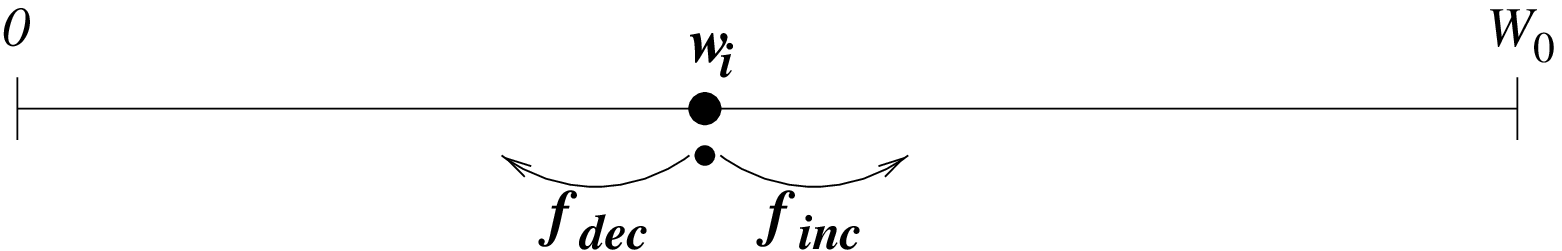}~\\[4mm]
Figure~3: One-dimensional random walk.
\end{center}
\medskip

Note that
the particle (i.e., the weight $w_i$) moves
to the left (resp., to the right)
with probability $1-1/n$ (resp., $1/n$).
The left (resp., right) end of the interval means that
the player $i$ becomes bankrupt (resp., a monopolist).
Thus,
these two ends are absorbing walls.

In a monopolist game under constrained rule with $n=2$,
we have $\finc=\cinc/2$ and $\fdec=\cinc/2$.
Hence,
the above random walk becomes standard one (see, e.g., \cite{feller:book68}),
and it is well-known that
the particle in such a standard random walk
goes to one of the absorbing walls in finite steps with probability 1.
This proves that $P_*=1$ when $n=2$.
Then by induction,
we can prove $P_*=1$ when $n>2$;
thus,
we have the following theorem.

\begin{theo}\label{theo:constraint}
Under constrained rule,
a monopolist emerges (in finite steps) with probability 1.
\end{theo}

\medskip\noindent
\underline{Local Rule}

\noindent
In constrained rule,
for computing $\fdec$,
we need the number of remaining players;
that is,
weights cannot be updated locally.
In general,
in order to be competitive,
an updating rule must not be local.
Thus,
to see the importance of competition,
we consider here the following purely local updating rule.
\begin{equation}
\finc~=~\cinc,{\rm~~and~~}\fdec~=~\cdec.
\end{equation}

Notice that
for this local rule (and the next semi local rule),
the notion of monopolist is less clear than constrained rule,
because the notion of ``enough amount of money'' is not clear.
Here we simply consider it as $W_0/2$,
a half of the total initial weight.
That is,
we regard a single surviver as a monopolist
if his weight is more than $W_0/2$;
hence,
$P_*$ is the probability that
the game reaches to the state
where $w_{i_1}\ge W_0/2$ for some $i_1$
and $w_i=0$ for the other $i$.

We first discuss one feature of this updating rule.
In the following,
let us fix $\cdec=1$.
Our computer experiments show that
the probability to have
a single surviver (in a reasonable amount of steps)
drops rapidly when $\cinc\ge n+1$.
The reason is clear from the following fact.

\begin{theo}\label{theo:local_exp}
Fix $\cdec$ to be one,
and consider one player's weight.
For any $t$,
it increases,
by
$\displaystyle t\Bigl(\frac{\cinc}{n}-1\Bigr)$
on average, after $t$ steps.
\end{theo}

Thus,
if $\cinc>n$,
then it is quite likely that all players increase their weights,
and thus no bankrupt appears in the game.
On the other hand,
if $\cinc<n$,
then every player dies quickly,
and hence,
no monopolist occurs
even though someone may become the last player.
This means that
the most crucial case is the case $\cinc=n$.
Next we discuss $P_*$ for such a case.

Recall that
$P_*$ is the probability that,
at some point in the game,
all but one become bankrupt
and that the survived player has weight $\ge W_0/2$.
Since it is difficult to estimate $P_*$ directly,
we analyze the following probability $P'_*$ instead of $P_*$:
$P'_*$ is the probability
that at least one player's weight reaches to $W_0/2$
and no more two players have weight
larger than sufficiently large value, say, $kW_0$ for some $k>0$.
Notice that
if a monopolist emerges at some point,
then clearly,
someone needs to reach $W_0/2$ in the game.
Furthermore,
it is unlikely that
two players reach to $kW_0$ and one of them become bankrupt afterwards.
Thus,
we may regard $P'_*$ as an upper bound of $P_*$.
For this $P'_*$,
we have the following bound.

\begin{theo}\label{t:local_prob}
For any $W_1$ and sufficiently large $W_2$,
we have

\[
P'_{*}
~<~
\Bigl(1 - \frac{I}{W_2+n}\Bigr)^{n}
+ \frac{nI}{W_2}\Bigl(1-\frac{I}{W_2+n}\Bigr)^{n-1}
- \Bigl(1-\frac{I}{W_1}\Bigr)^{n}.
\]
\end{theo}

For example,
by taking $W_1$ and $W_2$ as $\frac{nI}{2}$ and $knI$ respectively,
we have

\[
P'_{*}
~<~e^{-I/(kI+1)}+e^{-I(n-1)/(kI+1)n}/k-e^{-2}
~\approx~(1+1/k)e^{-1/k}-e^{-2},
\]

\noindent
which is less than 0.6 if $k=1$.
On the other hand,
our computer experiments show that
$P_*$ is less than 0.5 for various sets of parameters.

\medskip\noindent
\underline{Semi Local Rule}

\noindent
As a third updating rule,
we consider somewhat mixture of the above two rules.
It keeps a certain amount of locality,
but it still has some constraint.
This rule is defined as follows.
\begin{equation}
\finc~=~\min(\cinc,W_0-\sum_j w_j),{\rm~~and~~}\fdec~=~\cdec.
\end{equation}
That is,
we want to keep the total weight smaller than $W_0$,
where $W_0$ is the total initial weight.
Thus,
a winner can gain $\cinc$ (in net, $\cinc-\cdec$)
if there is some room to increase its weight.
In this case,
only the winner needs to know the current total weight,
or the amount of room to the limit $W_0$,
and the other players can update its weight locally.

Our computer experiments show that
the probability $P_*$ that a monopolist emerges is fairly large
if $\cinc$ is large enough, say $\cinc\ge 2n$.
On the other hand,
$P_*$ gets small when $\cinc$ is small,
which is explained in the same way as local rule.
Although we have not been able to prove that
$P_*$ is large for sufficiently large $\cinc$,
we can give some analytical result supporting it.

Here instead of analyzing $P_*$,
we estimate (i) the average number of steps
until all but one players become bankrupt,
and (ii) the average number of steps until
the total weight (which is initially $W_0$) becomes $W_0/2$.
Let $\Stoone$ and $\Stohalf$ denote
the former and the latter numbers respectively.
We prove below that
$\Stoone$ is smaller than $\Stohalf$
if $W_0$ is large enough.
This means that
it is likely that
at the time when all but one players become bankrupt,
the total weight,
which is the same as the survivor's weight,
is larger than $W_0/2$,
that is,
the surviver is a monopolist.

\begin{theo}\label{theo:semilocal}
Fix again $\cdec$ to be one.
If $I$ $\ge$ $(\ln6)n(n-2)$ and $\cinc\ge 2n$,
then we have
$\Stohalf$ $>$ $\Stoone$.
\end{theo}

\smallsection{Efficiency Analysis}

In this section we discuss how fast a monopolist emerges in games with
constrained rule.  We estimate an upper bound on the average number of
steps needed for monopoly to emerge, and we give some justification
(not a rigorous proof) supporting that it is $O(n^2(In/\cinc)^2)$.

We start with some definitions and notations that are used through the
section.  Here we modify our monopolist game and define a variant of
monopolist game.  Let $\gamezero$  denote the original monopolist
game.  We will denote by $\gameone$ a variant of $\gamezero$ in which
no bankrupt player can win.  Thus, in $\gameone$, the winning
probability of remaining players is $1/n'$ instead of $1/n$. As we
will see $\gameone$ is useful for induction and it is easier to
analyze.

These two game types are defined on different probability spaces.  Let
us define them more precisely.  For all two game types, (the execution
of) a game is specified by a {\em game sequence}, i.e., a string from
$\onenstar$ that defines a history of winners.  (Precisely speaking,
we also need to consider infinite strings; but as we see below, we may
ignore infinite strings.)  We say that a game sequence $x$ {\em kills}
a player $i$ if $w_i$ becomes 0 (or, negative) in the game following
$x$ just after the end of $x$, and we say that $x$ {\em derives a
monopolist} if the second last player is killed and monopoly emerges
just after $x$.  We say that a game sequence $x$ is {\em valid}
(resp., {\em strongly valid}) if it derives a monopolist and no prefix
of it derives a monopolist (resp., $x$ contains no indices of
previously killed players).  Note that the meaning of these notions
may vary depending on game types.  Now for any $n$, let $\domzero$
(resp., $\domone$) be the set of game sequences for $n$ player games
that are strongly valid w.r.t.\ $\gamezero$ (resp., valid w.r.t.\
$\gameone$).  For each $x$ in $\domzero$, its probability $\Pr{x}$ is
$n^{-|x|}$.  On the other hand, the probability $\Prone{y}$ of
$y\in\domone$ depends on the number of remaining players, and it is
rather complicated.  (We omit specifying $\Prone{y}$ because it is not
important for our discussion.)  Note that $\domzero$ and $\domone$ are
all prefix free.  Also it is not hard to show that $\Pr{\domzero}$ and
$\Prone{\domone}$ are one.  (For example, $\Pr{\domzero}=1$ follows
from Theorem~\ref{theo:constraint}.)  Therefore, we may regard
$\domzero$ and $\domone$ as a probability space of the corresponding
game, and we do not have to worry about infinite strings.

We denote by $T(n,I_1,\ldots,I_n)$ (resp., $\Tone(n,I_1,\ldots,I_n)$)
the number of steps needed until monopoly emerges in $\gamezero$
(resp, $\gameone$) with $n$ players and initial weight
$I_1,\ldots,I_n$.  When all the weights are equal, we use the simpler
notation $T(n,I)$.  Our goal is to get some upper bound on
$\Exp[T(n,I)]$.  But instead, we will analyze an upper bound on
$\Exp[\Tone(n,I)]$, which gives us an upper bound on $\Exp[T(n,I)]$,
as the following lemma guarantees.  (The proof is intuitively clear
and it is omitted in this abstract.)

\begin{lemm}\label{lemm:eff_zerovsone}
There exists $c_1$ such that
for any sufficiently large $n$ and any $I$,
we have $\Exp[T(n,I)]$ $\le$ $c_1n\Exp[\Tone(n,I)]$.
\end{lemm}

Now we analyze the convergence speed of $\gameone$.  For our analysis,
we split a game execution into stages where each stage is a part of
the game until some amount of players become bankrupt.  More
specifically, we denote by $\tone(n,I_1,\ldots,I_n)$ the number of
steps needed in a game with $n$ players and initial weights
$I_1,\ldots,I_n$ until at least $1$ player becomes bankrupt.  The
following lemma relates the two terms $\Tone(n,I_1,\ldots,I_n)$ and
$\tone(n,I_1,\ldots,I_n)$.

\begin{lemm}\label{lemm:eff_rec}
For any $n$ and $I_1,\ldots,I_n$, there exists a constant $c_2$,
$c_2\geq 1$, and weights $I'_1,\ldots,I'_{n-c_2}$ such that the
following inequality holds.

\[
\Exp[\Tone(n,I_1,\ldots,I_n)]
~\leq~\Exp[\tone(n,I_1,\ldots,I_n)]+\Exp[\Tone(n-c_2,I'_1,\ldots,I'_{n-c_2})].
\]
\end{lemm}

By this lemma we can use induction for bounding the expected value of
$\Tone$. Recall that when analyzing $\tone(n,I_1,\ldots,I_n)$, by the
way it is defined, no player becomes bankrupt, and thus, the amount of
decrement is fixed to $\cinc/n$.  Thus, $\gameone$ until at least one
player becomes bankrupt is regarded as a $n$-dimensional random walk,
which is much easier to analyze.  In fact, we can use the following
lemma.

\begin{lemm}\label{lemm:eff_anticb}
Let $X$ be a random variable that is 1 with probability $1/n$ and 0
with probability $1-1/n$, and let $S = X_1+\ldots+X_t$, the sum of the
outcomes of $t$ random trials of $X$.  Then, for some constant
$\alpha>0$, the following holds for any $t$ and $n$.

\[
{\rm Pr}\left\{
S\le{t\over n}-\alpha\sqrt{{t\over n}}
\right\}
~>~1/3.
\]
\end{lemm}

Now we are now ready to make the following claim\footnote{%
We do not have a rigorous proof for this result,
and for this reason we stated it as a claim.}.

\beginsome{Claim.}
\[
\Exp[T(n,I)]~=~O\left(n^2\left({In\over\cinc}\right)^2\right).
\]
\endsome

\beginsome{Justification.~}
We start with estimating $\Exp[\tone(n,I_1,\ldots,I_n)]$ by using the
above lemma.  For a given $t$ and for any $i$, Let $t_i$ be the number
of times that player $i$ wins within $t$ steps.  Then $w_i$, the
weight of player $i$, in $\gameone$ is expressed as follows.

\[
w_i
~=~
I_i+\cinc t_i-\cdec t
~\le~
I_i+\cinc\left(t_i-{t\over n}\right).
\]

\noindent
(For simplifying our notation, we use $c$ to denote $\cinc$ in the following.)

Moreover, since $\gameone$ until at least one player becomes bankrupt
is regarded as a $n$-dimensional random walk, we can use
Lemma~\ref{lemm:eff_anticb} to show that the following event happens
with probability bigger than $1/3$.

\[
w_i
~=~
I_i+c\left(t_i-{t\over n}\right)
~\le~
I_i+c\left({t\over n}-\alpha\sqrt{{ t\over n}}-{t\over n}\right)
~=~
I_i-c\alpha\sqrt{{t\over n}}.
\]

Therefore, with probability more than $1/3$, the weight of player $i$
becomes zero or negative if $c\alpha\sqrt{t/n}$ $\ge$ $I_i$, that is,
$t$ $\ge$ $(I_i/c\alpha)^2n$.  Now sort players by their initial
weights, and define $P$ to be the set of the first (i.e., the
smallest) $n/2$ players.  Since the total weight is $W_0$ (at any
step), all players in $P$ have weight at most $2W_0/n$ and therefore,

\[
{\rm Pr}\{\,
\mbox{$w_i\le 0$ in
$t_0$ $=$ $\displaystyle n\left({2W_0\over nc\alpha}\right)^2$ steps}
\,|\,i\in P\,\}~>~{1\over 3}.
\]

Moreover, {\em if} we can assume that each player in $P$ become
bankrupt independently, we also have the following probability: 

\[
{\rm Pr}\{\,
\mbox{There exists $i\in P$, such that $w_i< 0$ in
$t_0$ steps}
\,\}~ > ~{1-\rpr{{2\over 3}}^{n/2}}.
\]

From this observation, it is reasonable to bound
$\Exp[\ttwo(n,I_1,\ldots,I_n)]$ by $c_3t_0$ for some constant $c_3$
since for most of the valid game sequences (a ${1-\rpr{{2/3}}^{n/2}}$ fraction of them) this bound holds.

Now combining the above lemmas and the obtained bound,
we have

\[
\begin{array}{lcl} 
\Exp[\Tone(n,I)]
&\le&\Exp[\tone(n,I_1,\ldots,I_n)]+\Exp[\Tone(n-c_2,I'_1,\ldots,I'_{n-c_2})]\\
&\le&
\displaystyle
c_3n\left({2W_0\over{nc\alpha}}\right)^2+\Exp[\Tone(n-c_2,I'_1,\ldots,I'_{n-c_2})]\\
&\le&
\displaystyle
c_3n\left({2W_0\over{nc\alpha}}\right)^2
+c_3(n -1)\left({2W_0\over{(n-1)c\alpha}}\right)^2
+\Exp[\Tone(n-c'_2,I'_1,\ldots,I'_{n-c'_2})]~\cdots\\
&\le&
\displaystyle
c_3n\left({W_0\over c\alpha}\right)^2 \leq c_4n\left({In\over c}\right)^2,
~~~~\mbox{for some constant $c_4$.}
\end{array}
\]

\noindent 
From this and Lemma~\ref{lemm:eff_zerovsone},
we obtain the desired bound.
\endproof


\OMIT{for submission version
\section*{Acknowledgements}
This research has been started
while the first author visited to CRM,
Centre de Recerca Matem\`{a}tica,
Insitut D'Estudios Catalans.
He express his sincere thanks
to CRM and Professor Josep D\'\i\/az for inviting him to CRM.
We thank many researchers;
in particular,
to Dr.\ Tanaka at the Institute of Physical and Chemical Research (RIKEN)
and Dr.\ Miyashita at NEC Fundamental Research Laboratories
for their kind guidance to this field,
and to Jose Balc\'azar, Miklos Santha, and Carme Torras
for their interest and discussion.
Do not forget Paul, Mase-sensei.}

\eject
\bigparagraph{Appendix}

Here we state the proofs of lemmas
and theorems not stated in the body of the paper.

\paragraphskip\noindent
\underline{Section 3}

We begin with Theorem~\ref{theo:local_exp}.
To prove it,
we first estimate,
the probability $P_{I,W}$ that
one player, say player 1,
whose initial weight is $I$
obtains weight $W$ at some point in the game.
By using results in random work~\cite{feller:book68},
we can easily analyze this probability and obtain the following bounds.

\begin{fact}\label{fact:local_oneplayer}
For any $I$ and $W$,
we have
$\displaystyle
{I\over{W+n}}~\le~P_{I,W}~\le~{I\over W}$.
\end{fact}

\beginsome{Theorem~\ref{theo:local_exp}}
Fix $\cdec$ to be one,
and consider one player's weight.
For any $t$,
it increases,
by
$\displaystyle t\Bigl(\frac{\cinc}{n}-1\Bigr)$
on average, after $t$ steps.
\endsome

\beginproof
Consider $P'_{*}$
with $W_1$ and sufficiently large $W_2$.
Note first that

\[
\begin{array}{lcl}
\Prm{less than two players' weight reach to $W_2$}
&=&
\displaystyle
(1-P_{I,W_2})^{n}+nP_{I,W_2}(1-P_{I,W_2})^{n-1}\\
\Prm{no player's wealth reaches to $W_1$}
&=&
\displaystyle
(1-P_{I,W_1})^{n}.
\end{array}
\]

Then by definition of $P'_{*}$,
we have

\[
\begin{array}{lcl}
P'_{*}
&=&
\displaystyle
(1-P_{I,W_2})^{n}+nP_{I,W_2}(1-P_{I,W_2})^{n-1}-(1-P_{I,W_1})^{n}\\
&<&
\displaystyle
\Bigl(1-\frac{I}{W_2+n}\Bigr)^{n}
      +\frac{nI}{W_2}\Bigl(1-\frac{I}{W_2+n}\Bigr)^{n-1}
      -\Bigl(1-\frac{I}{W_1}\Bigr)^{n}.
\end{array}
\]
\square

Next we prove Theorem~\ref{theo:semilocal}.
In the following,
we only consider games with semi local rule specified in the theorem;
that is,
the game starts with $n$ players,
the bound is the same as the total initial weight $W_0=nI$,
and $\cinc\ge2n$.

Below,
for a given $k$ $\le$ $n$,
we consider the situation that $k$ players are left,
and analyze their total weight and the weight of the poorest player.
For this,
we use random walks representing respectively,
the total weight,
{\em a random walk for the total},
and the weight of currently the poorest player,
{\em a random walk for the poorest}.
These random walks are expressed as follows.

\displayskip
\begin{center}
\psbox[width=0.6\textwidth]{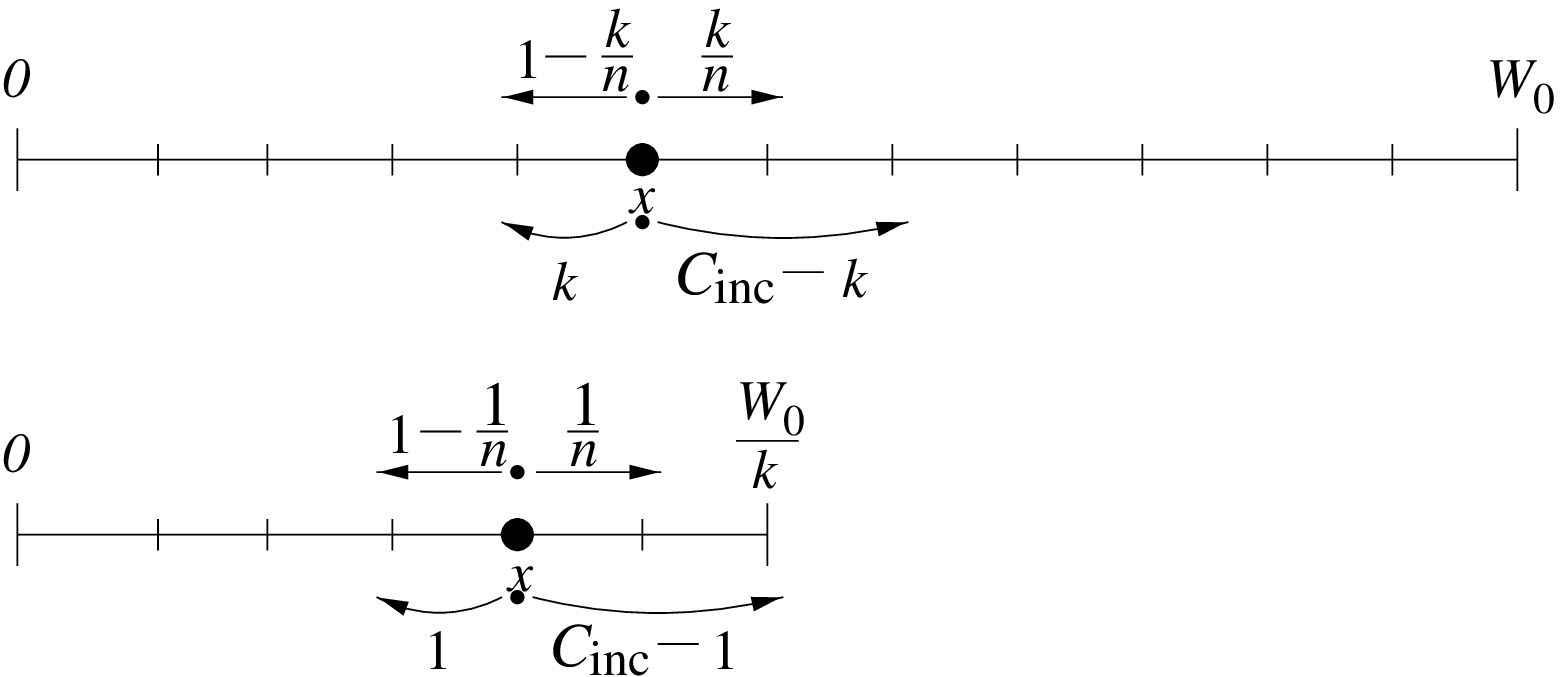}~\\[4mm]
Figure~4: Random walk for the total (above) and for the poorest (below)
\end{center}
\medskip

Since the total weight is at most $W_0$,
the poorest player cannot have more than $W_0/k$.
Also as soon as
the poorest player obtains $W_0/k$,
he cannot be the poorest,
and he (or, his role) is replaced with some currently poorest player,
whose weight is again less than $W_0/k$.
In any case,
if we only consider the weight of the poorest player,
we can assume that
the rightmost end is a reflecting wall.
On the other hand,
by definition of semi local rule,
the rightmost end of a random walk for the total is a reflecting wall.
Thus,
in both random walks,
the rightmost walls are reflecting walls;
hence,
the particles are eventually absorbed in the leftmost walls.
Here we discuss the difference
between the number of steps until the particles are absorbed.

\beginsome{Theorem~\ref{theo:semilocal}}
Fix again $\cdec$ to be one.
If $I$ $\ge$ $(\ln6)n(n-2)$ and $\cinc\ge 2n$,
then we have
$\Stohalf$ $>$ $\Stoone$.
\endsome

\beginproof
We first give an upper bound to $\Stoone$,
the average number of steps until all but one players become bankrupt.
For any $k\le n$,
consider the situation in the game that $k$ players are left,
and let $\SP{k}$ be the average number of steps
that some player becomes bankrupt
from this situation.

By Lemma~A.1 below,
we have

\[
\begin{array}{l}
\displaystyle
\Stoone
~\le~
\sum_{k=1}^{n}\SP{k}\\
\displaystyle
\le~
n(e^{W_0/2n}+e^{W_0/3n}+\ldots+e^{W_0/(n^2)})
~\le~
ne^{W_0/2n}+n(n-2)e^{W_0/3n}
~\le~
2ne^{W_0/2},\\
\end{array}
\]

\noindent
where the last inequality holds
from our assumption $W_0$ $\ge$ $n(n-2)\ln 6$.

Next consider $\Stohalf$,
the average number of steps needed until
the total weight becomes $W_0/2$ from $W_0$.
Let $\ST{k,x,y}$ be
the average number of steps
until the total weight becomes $y$ from $x$
under the condition that none of remaining $k$ players become bankrupt.
Note first that
for any $k$ and $k'$ such that $k<k'$,
we have
$\ST{k,x,y}$ $\le$ $\ST{k',x,y}$.
Then, for any $x_0,...,x_{n-3}$,
we have

\[
\ST{n,W_0,x_0}+\ST{n-1,x_0,x_1}+\cdots+\ST{2,x_{n-3},W_0/2}
~\ge~
\ST{2,W_0,W_0/2}.
\]

\noindent
Hence,
$\Stohalf$ $\ge$ $\ST{2,W_0,W_0/2}$.
On the other hand,
$\ST{2,W_0,W_0/2}$ is bounded in the following way
by using Lemma~A.2 below
and the assumption that $\cinc\ge2n$.

\[
\begin{array}{lcl}
\ST{2,W_0,W_0/2}
&>&
\displaystyle
\frac{n-2}{2}
\Bigl(\Bigl(1+\frac{2}{n-2}\Bigr)^{\frac{W_0}{2}}-1\Bigr)
\Bigl(1 - e^{-2}\Bigr)\\
&\ge&
\displaystyle
\frac{n-2}{2}\cdot
\Bigl(e^{\frac{W_0}{2(n-2)}}\bigl(1-e^{-2}\bigr)\Bigr)
~\ge~
\frac{n-2}{3}\cdot e^{W_0/2(n-2)}.
\end{array}
\]

Comparing both bounds,
we have $\Stohalf$ $\ge$ $\Stoone$
if $W_0$ $\ge$ $n(n-2)\ln 6$.
\endproof

\beginsome{Lemma A.1.}
Consider the situation in the game
that $k$ players are left,
and let $\SP{k}$ be
the average number of steps
that some player becomes bankrupt from this situation.
Then we have
$\SP{k}$ $\le$
$\displaystyle{ne^{\frac{W_0}{nk}}}$.
\endsome

\beginproof
Consider a random walk for the poorest.
As explained above,
we may assume that
the random walk has a reflecting wall
at $W_0/k$ and an absorbing wall at 0.
Here for showing an upper bound,
we assume,
as the worst case,
that the initial weight of the poorest is $W_0/k$;
that is,
the random walk starts off at $W_0/k$.
Then $\SP{k}$ is bounded by
the average number of steps
until the particle reaches to the absorbing wall at 0.

Let $t_{x}$ be the random variable denoting
the number of steps
starting off at $x$ and arriving at $x-1$ for the first time.
Then we can evaluate $t_{x}$ by

\[
t_{x}~=~\left\{
\begin{array}{ll}
1,&
\mbox{with probability $1-1/n$, and}\\
1+t_{x+\cinc-1}+\ldots+t_{x+1}+t_{x},
&\mbox{otherwise.}
\end{array}\right.
\]

Define $e_{x}$ to be the expectation of $t_{x}$.
Then we have

\[
e_{x}~=~
\frac{n}{n-1}+\frac{1}{n-1}\bigl(e_{x+\cinc-1}+\ldots+e_{x+1}\bigr).
\]

Then by induction on $x$,
we can show
that $e_{W_0/k-x}$ $\le$ $\bigl(1+1/(n-1)\bigr)^{x}$.
Thus,
we have

\[
\SP{k}~\le~\sum_{x=0}^{W_0/k-1}e_{W_0/k-x}
      ~\le~(n-1)\Bigl(1+\frac{1}{n-1}\Bigr)^{{W_0\over k}}
      ~\approx~(n-1)e^{\frac{W_0}{nk}}.
\]
\square

\beginsome{Lemma A.2.}
For any $x$ and $y$, $x>y$,
we have
\[
\ST{2,x,y}~>~
{\frac{n-2}{2}
\Bigl(\Bigl(1+\frac{2}{n-2}\Bigr)^{\frac{W_0-y}{2}}
-\Bigl(1+\frac{2}{n-2}\Bigr)^{\frac{W_0-x}{2}}\Bigr)
(1 - e^{-2})}.
\]
\endsome

\beginproof
Consider a random walk for the total (see Figure~4).
By dividing all parameters by $k=2$,
we can modify it to essentially the same random walks as Figure~5.

\displayskip
\begin{center}
\psbox[width=0.6\textwidth]{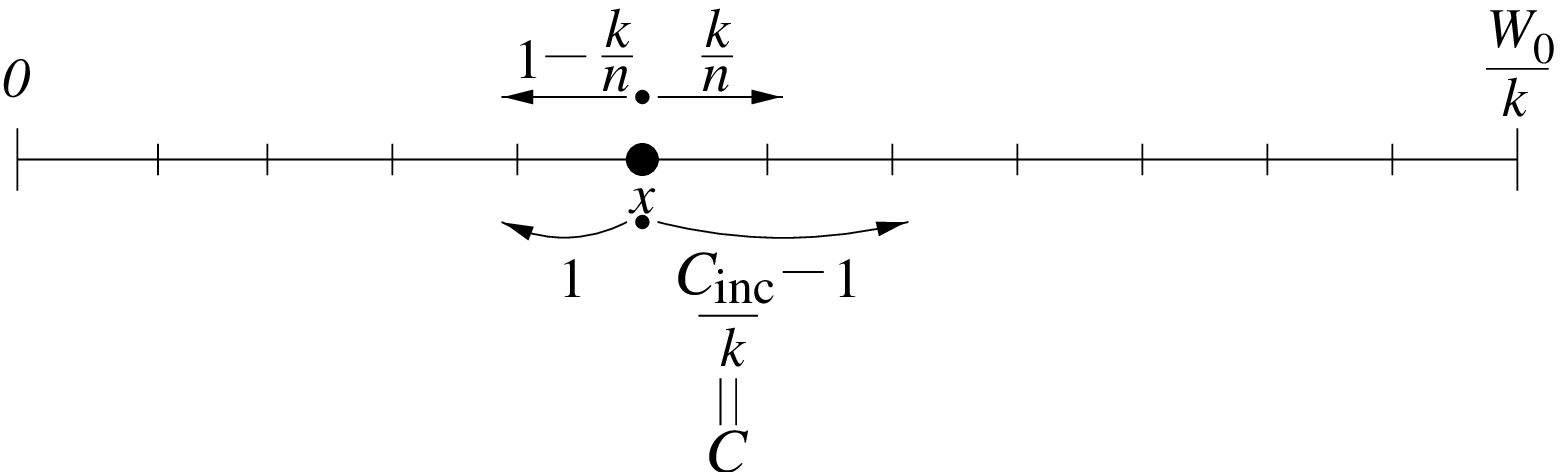}~\\[4mm]
Figure~5:
A modification of a random walk for the total
\end{center}
\medskip

Note that this random walk
is quite similar to a random walk for the poorest (see Figure~4 below).
Thus,
for $e_{x}$,
the average number of steps
starting off at $x$ and arriving at $x-1$ for the first time,
a similar argument derives that the following relation
(here $c$ $=$ $\cinc/2$).

\[
e_{x}~>~1 + \frac{2}{n-2}
        \bigl(e_{x+c-1} + \ldots + e_{x+1}\bigr).
\]

Then it is not so hard to see that

\[
e_{W_0/2-x}
~\ge~
\Bigl(1+\frac{2}{n-2}\Bigl)^{x}
-\Bigl(1+\frac{2}{n-2}\Bigr)^{x-c+1}\\
~=~
\Bigl(1+\frac{2}{n-2}\Bigl)^{x}
\Bigl(1 - \Bigl(1+\frac{2}{n-2}\Bigr)^{-c+1}\Bigr).
\]

\noindent
Now by using our assumption that $c$ $=$ $\cinc/2$ $\ge$ $n$,
we can derive the following bound.

\[
\begin{array}{lcl}
\ST{2,x,y}
&\ge&
\displaystyle
\sum_{x=(W_0-x)/2}^{(W_0-y)/2-1}
\Bigl(1+\frac{2}{n-2}\Bigl)^{x}
\Bigl(1-\Bigl(1+\frac{2}{n-2}\Bigr)^{-c+1}\Bigr)\\
&\ge&
\displaystyle
\frac{n-2}{2}
\Bigl(\Bigl(1+\frac{2}{n-2}\Bigr)^{\frac{W_0-y}{2}}
-\Bigl(1+\frac{2}{n-2}\Bigr)^{\frac{W_0-x}{2}}\Bigr)
(1 - e^{-2}).
\end{array}
\]
\square

\paragraphskip\noindent
\underline{Section 4}

\beginsome{Lemma~\ref{lemm:eff_rec}}
For any $n$ and $I_1,\ldots,I_n$,
there exists a constant $c_2$, $c_2\geq 1$,
and weights $I'_1,\ldots,I'_{n-c_2}$ such that the following inequality holds.

\[
\Exp[\Tone(n,I_1,\ldots,I_n)]
~\leq~\Exp[\tone(n,I_1,\ldots,I_n)]+\Exp[\Tone(n-c_2,I'_1,\ldots,I'_{n-c_2})].
\]
\endsome

\beginproof
Let $Y\subseteq\domone$ be the set of all valid game sequences $y$
such that the number of players becomes strictly smaller than $n$ for the first
time just after $y$.  By definition of
$\Exp[\Tone(n,I_1,\ldots,I_n)]$, we have the following equality.

\[
\displaystyle
\Exp[\Tone(n,I_1,\ldots,I_n)]
~=~
\sum_{x\in\domone}\Prone{x}\cdot|x|
~=~
\sum_{x\in\domone, y\in Y\atop x=yz}\Prone{yz}(|y|+|z|).
\]

Notice here that we can split $\Prone{yz}$ in two factors, $\Prone{y}$
and $\Proney{z}$, where $\Proney{z}$ determines the probability of $z$
after the game follows $y$.  Also note that the set $Y_y$ of strongly
valid $z$ depends on $y$.  Thus, we can rewrite the above expression
as follows.

\[
\begin{array}{l}
\Exp[\Tone(n,I_1,\ldots,I_n)]\\
~=~
\displaystyle
\sum_{y\in Y}\sum_{z\in Y_z}\Prone{y}\Proney{z}\cdot|y|
+\sum_{y\in Y}\sum_{z\in Y_z}\Prone{y}\Proney{z}\cdot|z|\\
~=~
\displaystyle
\left(\sum_{y\in Y}\Prone{y}|y|\right)\sum_{z\in Y_z}\Proney{z}
+\sum_{y\in Y}\Prone{y}\left(\sum_{z\in Y_z}\Proney{z}\cdot|z|\right)\\
~=~
\displaystyle
\Exp[\tone(n,I_1,...,I_n)]\sum_{z\in Y_z}\Proney{z}
+\sum_{y\in Y}\Prone{y}\Exp[\Tone(n_y,I_{i_1(y)},\ldots,I_{i_{n_y}(y)})]\\
~\le~
\displaystyle
\Exp[\tone(n,I_1,\ldots,I_n)]
+\Exp[\Tone(n-c',I'_1,\ldots,I'_{n-c'})].
\end{array}
\]

\noindent
where the values of $n_y$ and $I_{i_1(y)},\ldots,I_{i_{n_y}(y)}$ are
determined by the result of the game following $y$.  On the other
hand, $c'$ and $I'_i$ are chosen so that the value of
$\Exp[\Tone(n_y,I_{i_1(y)},\ldots,I_{i_{n_y}}(y))]$ is maximized.
These values always exist since even if there is an infinite number of
game sequences $y$ that appear on the summation, there is only a
finite number of possible values for $n_y$ (since $n_y$ must be
between $1$ and $n-1$) and $I_{i_j(y)}$ (since $\sum_j
I_{i_j(y)}=In$).
\endproof

\beginsome{Lemma~\ref{lemm:eff_anticb}.}
Let $X$ be a random variable
that is 1 with probability $1/n$
and 0 with probability $1-1/n$,
and let $S = X_1+\ldots+X_t$,
the sum of the outcomes of $t$ random trials of $X$.
Then, for some constant $\alpha>0$,
the following holds for any $t$ and $n$.

\[
{\rm Pr}\left\{
S\le{t\over n}-\alpha\sqrt{{t\over n}}
\right\}
~>~1/3.
\]
\endsome

\beginproof
We estimate the probability
that the statement of the lemma is false
and show that it is less than 2/3.
That is,
we upper bound the following probability.

\[
{\rm Pr}\left\{
S>{t\over n}-\alpha\sqrt{{t\over n}}
\right\}
~=~
{\rm Pr}\left\{
S>{t\over n}
\right\}
+{\rm Pr}\left\{
{t\over n}\ge S>{t\over n}-\alpha\sqrt{{t\over n}}
\right\}.
\]

The first term in the sum is bounded by $1/2$;
see, e.g., \cite{stat_table:book72}.
Let $s$ be the smallest integer such that
$s>{t/n}-\alpha\sqrt{{t/n}}$.
We calculate,
by using Stirling's approximation,
the second term of the above sum as follows.

\[
\begin{array}{lcl}
\displaystyle
{\rm Pr}\left\{{t\over n}\ge S>{t\over n}-\alpha\sqrt{{t\over n}}\right\}
&=&
\displaystyle
\sum_{i = s}^{t/n}
{t\choose i}\left({1\over n}\right)^i\left(1 -{1\over n}\right)^{t-i}
~=~
\sum_{i=s}^{t/n}{t \choose i}{(n-1)^{t-i}\over n^t}\\
&\le&
\displaystyle
\sum_{i=s}^{t/n}\sqrt{{t\over 2\pi i(t-i)}}\left({t\over t-i}\right)^t
                \left({t-i\over i}\right)^i{(n-1)^{t-i}\over n^t}\\
&\le& 
\displaystyle
\sum_{i=s}^{t/n}
\sqrt{{t\over 2\pi i(t-i)}}\left({t(n-1)\over n(t-i)}\right)^t
                           \left({t-i\over i(n-1)}\right)^i.
\end{array}
\] 

\noindent 
Also routine calculations show that
$\left({t(n-1)\over n(t-i)}\right)^t\left({t-i\over i(n-1)}\right)^i$
is always less than $1$
for $s$ $\le$ $i$ $\le$ $t/n$
and that this factor is maximized when $i={t/n}$.
From this by simple calculation,
we obatain the desired bound
with $\alpha = \sqrt{\pi/12}$.

\OMIT{
\[
\begin{array}{lll}
\displaystyle {\rm Pr}\spr{ {t\over n} \geq S > {t\over n} -\alpha\sqrt{{t\over n}}}  
&\leq &  
\displaystyle \sum_{i=s}^{t/n} \sqrt{{t\over 2\Pi i(t-i)}} \\
&\leq & 
\displaystyle \alpha\sqrt{{t\over n}}\sqrt{ {t\over 2\Pi(t/2n)(t-(t/2n))}} \\
& = & 
\displaystyle \alpha\sqrt{{t\over n}}\sqrt{ {2n^2 \over \Pi t(2n-1)}} \\
&\leq & 
\displaystyle \alpha \sqrt{{2\over \Pi}}
\end{array}
\]}

\end{document}